\documentclass{article}
\usepackage{ijcai25}

\usepackage{xr}
\usepackage{times}
\usepackage{soul}
\usepackage{url}
\usepackage[hidelinks]{hyperref}
\usepackage[utf8]{inputenc}
\usepackage[small]{caption}
\usepackage{graphicx}
\usepackage{amsmath}
\usepackage{amsthm}
\usepackage{booktabs}
\usepackage{algorithm}
\usepackage{algorithmic}
\usepackage[switch]{lineno}
\usepackage{soul}
\usepackage{float}
\usepackage{mathtools}

\usepackage{xcolor}
\usepackage{amsfonts}

\def\ltlweaknext{\mathcal{W}}
\def\ltluntil{\mathcal{U}}
\def\ltlreleases{\mathcal{R}}

\def\ltlnext{\bigcirc}
\def\ltlfinally{\Diamond}
\def\ltlglobally{\Box}

\newcommand{\mathimg}[1]{\vcenter{\hbox{\includegraphics[height=2ex]{imgs/#1}}}}

\newcommand{\parafango}[1]{\vspace{-1mm}\paragraph{#1}}

\newcommand{\QUEST}[1]{\textcolor{magenta}{#1}}

\newcommand{\sm}[2]{\textcolor{orange}{#1} {\color{orange}[{\it #2}]}}
\newcommand{\smc}[1]{\textcolor{magenta}{#1}}
\newcommand{\lsl}[1]{\textcolor{blue}{#1}}
\newcommand{\ml}[1]{\textcolor{magenta}{#1}}

\title{A Neuro-Symbolic Framework for Sequence Classification\\with Relational and Temporal Knowledge}

\author{
Luca Salvatore Lorello$^{1,2}$\footnote{Part of the work was done while LSL was visiting KU Leuven. 
}
\and
Marco Lippi$^3$\and
Stefano Melacci$^{4}$
\affiliations
$^1$University of Pisa, $^2$University of Modena and Reggio Emilia,\\
$^3$University of Florence, $^4$University of Siena\\
\emails
luca.lorello@phd.unipi.it,
marco.lippi@unifi.it,
stefano.melacci@unisi.it
}

\begin{document}

\maketitle

\begin{abstract}
One of the goals of neuro-symbolic artificial intelligence is to exploit background knowledge to improve the performance of learning tasks. However, most of the existing frameworks focus on the simplified scenario where knowledge does not change over time and does not cover the temporal dimension. In this work we consider the much more challenging problem of knowledge-driven sequence classification where different portions of knowledge 
must be employed at different timesteps, and 
temporal relations are available. Our experimental evaluation compares multi-stage neuro-symbolic and neural-only architectures, and it is conducted on a newly-introduced benchmarking framework. Results demonstrate the challenging nature of this novel setting, and also highlight under-explored shortcomings of neuro-symbolic methods, representing a precious reference for future research.
\end{abstract}

\section{Introduction}



Sequence classification is a very well-known task in machine learning which can become very challenging when decisions must be made on top of complex or hard-to-learn features and without suitable priors, especially when safety, reliability and accountability are of paramount importance (autonomous driving~\cite{roesener2016scenario}, industrial control systems~\cite{chakraborty2022device}, medicine~\cite{ivaturi2021comprehensive}, and others).
The task can be also approached by symbolic methods
exploiting grounded frameworks, such as regular expressions~\cite{galassi2005learning} and automata induction~\cite{angluin1982inference}, or temporal logics satisfiability~\cite{rozier2007ltl}. However, they tend to struggle in the presence of noise~\cite{umili2024deepdfa}, and may be unsuitable for open-world settings where background knowledge or 
inductive priors are not available at all.
Neuro-symbolic artificial intelligence~\cite{besold2021neural} aims to get the best of both worlds, by merging neural and symbolic methods, as in DeepProbLog~\cite{manhaeve2018deepproblog}, or Logic Tensor Networks~\cite{badreddine2022logic}.
However, these methods fight to accommodate inconsistencies between simplified experimental settings (needed to speed-up computations) and real world scenarios.


Little has been done for
sequence classification in presence of ($i$) 
``relational'' and ($ii$) ``temporal'' knowledge. 
The former describes 
features within each single time-step 
while the latter models information along the temporal dimension.
%
{For example, consider a safety-critical system where several cameras monitor an environment, controlling the movements of automated guided vehicles. At each timestep, every vehicle has to satisfy some safety properties regarding its movements, for example to avoid collisions; in addition, at some checkpoints it is necessary to assess that some tasks have been executed in a specific order throughout the temporal sequence.}

Motivated by this challenging and still under-explored setting, this paper collects several contributions. 
($a$) We propose a new benchmarking framework for 
 sequence classification (and similar tasks) in the presence of domain knowledge that extends also over the temporal dimension. Our framework is capable of generating datasets with multi-channel sequences of arbitrary length, by sampling given image classification datasets according to user-defined temporal and relational 
 specifications.
($b$) We share with the scientific community ready-to-use datasets for six challenging tasks, as well as baseline performances for neural-only and modular neuro-symbolic architectures. Building on top of a neural  net processing the raw data, our neuro-symbolic architecture consists of an automata-based temporal reasoner stacked on top of a relational symbolic module, both representatives of state of the art approaches in their respective categories.
($c$) Our detailed experimental study highlights two under-investigated problems: 
specific methods for temporal reasoning, highly-effective with propositional inputs, struggle when extended to 
a relational setting, even when the interface is ``propositionalized'' by an upstream 
reasoner, and, conversely, relational neuro-symbolic methods present training instabilities when coupled with downstream (temporal) reasoning components.

\section{Background}

\paragraph{Linear temporal logic over finite traces}
A dynamic discrete system is a triple $\langle S, R, s_0\rangle$, where $S$ is a set of (possibly infinite) states, $R$ is a transition relation between states, often non-deterministic, and $s_0 \in S$ is the initial state. An execution {\it trace} is a (possibly infinite) sequence of states $(s_0, s_1, s_2, \dots)$, such that $R(s_{i-1}, s_i), \forall i > 0$.
Linear Temporal Logic (LTL)~\cite{pnueli1977temporal} is a formalism for temporal reasoning over infinite execution traces of dynamic discrete systems, capable of expressing properties related to the reachability of states within a trace. Operators in LTL allow to express different types of properties. Given a sub-formula or an atomic proposition, indicated with $\phi$ or $\psi$ in the following, the {\it next} operator $\ltlnext$ models immediate future reachability, 
 i.e., $\ltlnext \phi$ means that $\phi$ has to hold at the next state. Other operators model eventual reachability ({\it finally/eventually} operator, $\ltlfinally \phi$, i.e., $\phi$ has to hold at least once in the future), invariance ({\it globally} operator, $\ltlglobally \phi$, i.e., $\phi$ has to hold on the entire subsequent path) and conditional reachability (such as the binary operators {\it until}, $\phi \ltluntil \psi$, and {\it releases}, $\phi \ltlreleases \psi$). 
Over the years, the original formulation of LTL has been modified in multiple ways: LTL over finite traces (LTL$_f$)~\cite{de2013linear} is a restriction which preserves the same syntax of LTL\footnote{LTL$_f$, introduces an additional operator, weak next, $\ltlweaknext \phi$, 
to deal with last state semantics in finite traces.} but redefines its semantics, in order to be applied to execution traces of finite length. This restriction also has practical repercussions: while LTL can express $\omega$-languages recognized by B\"uchi automata, reasoning over LTL$_f$ can be reduced to regular languages, accepted by deterministic finite state automata (DFA).
In this work we convert LTL$_f$ formulas to DFAs, using the flloat\footnote{\url{https://github.com/whitemech/flloat}} library.

\parafango{Neuro-symbolic integration of logic formulas}
The main challenge of neuro-symbolic integration~\cite{besold2021neural} consists in providing an interface between two components: learning by means of neural networks, which requires representations in a continuous space, and reasoning, which often benefits from discrete representations of symbols.
A popular approach for the integration of logic knowledge, known as the model-theoretic approach~\cite{marra2024statistical}, is to relax truth assignments, in a way which extends Boolean algebra in a continuous and differentiable space.
Fuzzy logic~\cite{badreddine2022logic} extends interpretations in the continuous range $[0, 1]$ and replaces Boolean conjunction with a t-norm.\footnote{In propositional logic, every other operator is constructed by exploiting the definition $\neg p = 1 - p$ and equivalence axioms, for fuzzy first order logic systems, additional components are required.} The obtained expressions are equivalent to the original Boolean formula at boundary values, but allow differentiability by means of a progressive transition between truth values, which can constrain the learning procedure \cite{gnecco2015foundations}. Different choices of t-norms are possible, each characterized by different advantages and drawbacks.
%
An alternative framework for model-theoretic neuro-symbolic integration, overcoming some issues with differentiable fuzzy logics \cite{van2022analyzing}, is based on probabilistic inference. In this framework, with strong theoretical and computational foundations in statistical and relational artificial intelligence~\cite{raedt2016statistical}, Boolean propositions are seen as Bernoulli random variables and logic connectives are interpreted as set operators.
Weighted model counting (WMC) is a general framework for probabilistic inference. 
Algebraic model counting (AMC)~\cite{kimmig2017algebraic} extends WMC by replacing Boolean operators with elements of an algebraic semiring, allowing to solve a plethora of probabilistic tasks within a single framework. WMC and AMC are in general $\textsc{\#p-complete}$~\cite{chavira2008probabilistic}, however they become tractable when logic formulas possess a specific structure. Knowledge compilation~\cite{darwiche2002knowledge} amortizes execution time by converting input formulas into equivalent target normal forms, on top of which clausal inference (and possibly other classes of tasks) can be executed in polynomial time. 
In this context, we mention sd-DNNF, the smooth and decomposable deterministic negated normal form, a popular knowledge compilation target language, which guarantees correct model counting in polynomial time, also in the presence of neutral and disjoint sums.

\section{Related Works}

\paragraph{Neuro-symbolic temporal reasoning}

Neuro-symbolic temporal reasoning has been employed for temporal formula induction, approximate satisfiability and sequence classification driven by background knowledge.
\cite{camacho2019learning} address formula induction in a symbolic fashion, by building a vocabulary of subformulas which are converted to alternating finite automata, and then composed to discover the target formula from positive and negative examples.
On the other hand, \cite{walke2021learning} address formula induction by means of interpretable neural networks, proposing the use of specialized recurrent layers for sequence classification, which can collectively be interpreted as an LTL$_f$ formula after training, by extracting a truth table from discretized weights.
Finally, \cite{umili2024deepdfa} attempt to learn DFA transition matrices from examples, by discretizing a recurrent neural network, regularized during training  
to produce discrete activations.
Exact LTL satisfiability is $\textsc{pspace-complete}$. However, recently, several approaches based on neural networks have been proposed to approximate satisfiability in polynomial time: \cite{xie2021embedding} employ message passing graph neural networks to learn embeddings for a DFA equivalent to the target LTL$_f$ formula;
\cite{mukherjee2022octal} exploit graph isomorphism networks to perform approximate model checking;
\cite{luo2022checking,luo2024learning} use recursive neural networks to generate explanations as traces satisfying the given formulas.

\parafango{Neuro-symbolic finite state automata}

There is a strong link between inference in DFA and recurrent neural networks, as the former performs trace acceptance using a state transition matrix and the latter updates an internal state according to a deterministic function.
On the other hand, the discrete nature of DFAs is particularly amenable to be encoded by (sets of) Boolean formulas, which, in turn, can be subject to neuro-symbolic integration.
Starting from this consideration, \cite{umili2023grounding} exploit background knowledge in the form of LTL$_f$ formulas to learn symbolic labels from sequence labels, by performing distant supervision~\cite{manhaeve2018deepproblog}.
The transition matrix is converted into a disjunction of Horn clauses 
and the entire logic program is finally encoded into a Logic Tensor Network~\cite{badreddine2022logic}
that can be trained end-to-end on sequence labels, by means of binary cross-entropy or a proposed semantic loss penalizing discordance between ground truth labels and the observed final state. 
%
\cite{manginas2024nesya} highlight the shortcomings of this method, by identifying a general failure on more complex datasets, consisting of larger DFAs, longer temporal horizons or more complex transition guards, and pinpointing the cause on the fuzzy encoding.
In their extension, each transition guard is compiled into an sd-DNNF, and the automaton is given probabilistic semantics, by means of AMC on the probability semiring, largely improving scalability. 

\parafango{Relationships to this paper} We take inspiration both from \cite{umili2023grounding} and 
 \cite{manginas2024nesya}, however we allow transition labels to contain constraints (i.e., first order predicates of known semantics, applied to terms which represent objects in finite domains). This distinction places our work closer to symbolic automata~\cite{veanes2010symbolic,veanes2013applications,d2017power}
 , and it is characterized by additional challenges.
We address these challenges in a straight-forward manner, by decoupling reasoning in two components: 
relational and temporal inference.
In this way, temporal inference is reduced to the propositional case, 
decoupling the additional challenges provided by a first order setting, from those posed by more complex temporal behaviors.
From a neuro-symbolic perspective, our decoupling approach also provides benefits in terms of knowledge injection: as the semantics of states is defined over constraint satisfaction problems, 
background knowledge is richer and more structured (i.e., constraints act as relations between concepts, while propositional formalisms are limited to assignments).

\section{Sequence Classification with Relational and Temporal Knowledge}
\label{sec:seq}


\paragraph{Problem definition}

Let $\mathcal{S} = ( [ x_0^{t}, \dots, x_{N-1}^{t} ] )_{t=0}^{T-1}$ be a sequence of length $T$ that consists, for each time step $t$, of $N$ perceptual stimuli belonging to as many \textit{perceptual domains $\mathcal{X}_j$}: $x_j^{t} \in \mathcal{X}_j$. Each $x_j^{t}$ is associated to a symbol $y^{t}_j$ that belongs to a \textit{symbolic domain} $y^{t}_j \in \mathcal{Y}_j$. As an example, $\mathcal{X}_0$ and $\mathcal{X}_1$ could be the domain of MNIST digits~\cite{lecun1998mnist} and Fashion-MNIST articles~\cite{xiao2017fmnist}, respectively, whereas $\mathcal{Y}_0$ and $\mathcal{Y}_1$ the corresponding sets of (symbolic) classes. To avoid confusing the stimulus/domain index with the time index, we will sometimes use letter subscripts in place of numbers (e.g., $A$, $B$ in place of $0$, $1$) to refer to different stimuli, symbols, and domains (e.g., $x_{A}^t$, $x_{B}^t$, $y_{A}^t$, $y_{B}^t$, $\mathcal{X}_{A}$, $\mathcal{X}_{B}$, $\mathcal{Y}_A$, $\mathcal{Y}_B$, etc.).

We consider the problem  of learning a binary classifier $f\colon\mathcal{S} \mapsto \{0, 1\}$ which maps the input sequence to the positive class if and only if $\mathcal{S} \models \mathcal{T}$, where $\mathcal{T}$ is a temporal specification that consists in an LTL$_f$ formula whose atomic symbols are 
first order relations grounded over
\textit{symbolic domains} $\mathcal{Y}_j$. 
Intuitively, this means that the sequence classification task consists in assessing whether two sets of properties jointly hold or not: ($i$) instantaneous relations among the $N$ symbolic domains which may or may not hold for any $t$ (``relational''), and ($ii$) meta-relations about the validity of each instantaneous property over time (``temporal''). The reasoning process in such setting can be reduced to querying the validity of each instantaneous property to build a trace of truth values, and then checking whether the trace 
satisfies
 the temporal specification.
In order to preserve LTL$_f$ conciseness and readability, we encode 
``relational'' (i.e., instantaneous) properties as constraints over finite domains~\cite{nethercote2007minizinc}, which allow to express arithmetic, identity-based, ordering-based, lexicographic, etc., relations with well-codified, compact, and efficient to compute constructs, known as global constraints. To further aid readability, we collect 
($i$) relational 
properties into a set of constraints $\mathcal{C}$, separately from the ($ii$) ``temporal'' formula in  $\mathcal{F}$. It is important to note that, in our setting, it is the joint effect of the LTL$_f$ formula and the properties with which its symbols are grounded  (i.e., the complete background knowledge is $\mathcal{T} = \mathcal{C} \cup \mathcal{F}$), which acts as decision rule. 

As an example, let us consider two popular digit classification datasets, MNIST and SVHN~\cite{netzer2011reading}, where class names can be also interpreted as class indices. Suppose that we have a sequence $\mathcal{S}$ where, at each time instant, we are given $N=3$ stimuli $(x_A^t, x_B^t, x_C^t)$, $x_A^t \in \mathcal{X}_A$, $x_B^t \in \mathcal{X}_B$, with $\mathcal{X}_A = \mathcal{X}_B =$ MNIST digits, while $x_C^t \in \mathcal{X}_C =$ SVHN digits. Let us consider the case in which the $N$ perceptual stimuli are mapped to $N$ symbols $(y_A^t, y_B^t, y_C^t)$, belonging to domains $\mathcal{Y}_A = \mathcal{Y}_B = [0, 9]$, and $\mathcal{Y}_C = [2,8]$ (integer).
As 
``relational''
knowledge we assume that two properties can potentially hold: $y_A + y_B = y_C$ and $y_A \not= y_B \wedge y_B \not= y_C \wedge y_C \not= y_A$,\footnote{We use MiniZinc~\cite{nethercote2007minizinc} semantics, thus enumerations are cast to integers in lexicographic order, allowing expressions like $A + B = C$ also for non-numeric symbols.} where we compactly indicate the latter with $\texttt{\small all\_different}(y_A,y_B,y_C)$. We distinguish the behavior between odd and even time instants to introduce a form of ``temporal'' knowledge: in even timesteps the sum constraint is expected to hold, while $\texttt{\small all\_different}$ is expected to hold for odd steps. The goal is to learn to solve a binary classification problem, distinguishing knowledge-coherent sequences $\mathcal{S}^+$'s from incoherent ones, $\mathcal{S}^-$'s.
The properties of our problem can more formally and concisely be defined as the quadruple $\langle\mathcal{X}, \mathcal{Y}, \mathcal{C}, \mathcal{F}\rangle$ where $\mathcal{X}$ and $\mathcal{Y}$ are the unions of all the perceptive and symbolic domains, respectively; $\mathcal{C}$ is the set encoding the relational knowledge; $\mathcal{F}$ is the temporal formula. 
We have: 
\begin{align*}
 \mathcal{X}\colon\quad&\mathcal{X}_A, \mathcal{X}_B = \left\{\mathimg{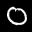}, \mathimg{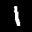}, \ldots, \mathimg{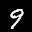}\right\}; \ \  
 \mathcal{X}_C = \left\{ \mathimg{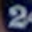}, \mathimg{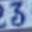}, \ldots, \mathimg{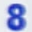}\right\}\\
 \mathcal{Y}\colon\quad&\mathcal{Y}_A, \mathcal{Y}_B = [0,9]; \quad\quad\quad\quad\ \ \ \ \hskip 0.5mm  
 \mathcal{Y}_C = [2,8]\\
 \mathcal{C}\colon\quad&p: y_A+y_B=y_C\\
 &q: \texttt{\small all\_different}(y_A,y_B,y_C)\\
 \mathcal{F}\colon\quad&p \wedge \ltlglobally (p \leftrightarrow \ltlnext q).
\end{align*}
Therefore, a sequence of triples such as $\mathcal{S}^+ = (\left[\mathimg{mnist0},\mathimg{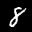}, \mathimg{svhn8}\right], \left[\mathimg{mnist1},\mathimg{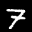}, \mathimg{svhn3}\right], \left[\mathimg{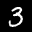},\mathimg{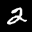}, \mathimg{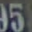}\right], \left[\mathimg{mnist2},\mathimg{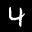}, \mathimg{svhn8}\right])$, corresponding to the trace $(\left[p, \neg q\right], \left[\neg p, q\right], \left[p, q\right], \left[\neg p, q\right])$, 
will be accepted by a knowledge-driven sequence classifier, while $\mathcal{S}^- = ([\mathimg{mnist0},\mathimg{mnist8}, \mathimg{svhn8}], [\mathimg{mnist1},\mathimg{mnist7}, \mathimg{svhn3}], [\mathimg{mnist3},\mathimg{mnist1}, \mathimg{svhn5}], [\mathimg{mnist2},\mathimg{mnist4}, \mathimg{svhn8}])$, corresponding to the trace $([p, \neg q], [\neg p, q], [\mathbf{\neg p}, q], [\neg p, q])$ will not, as $\mathbf{\neg p}$ (bold) violates the temporal property. 
\parafango{The LTLZinc Framework}
LTLZinc\footnote{
\url{https://github.com/continual-nesy/LTLZinc}} is a benchmarking framework for temporal reasoning tasks, capable of generating sequences from user-defined constraint specifications, involving relational and temporal knowledge.
LTLZinc receives as input the 
quadruple $\langle\mathcal{X}, \mathcal{Y}, \mathcal{C}, \mathcal{F}\rangle$, as well as additional options, and produces an output depending on either of two modes: ($i$) sequential and ($ii$) incremental.
In ($i$) \emph{sequential} (which could informally be categorized as ``learning about time''), the output will be a dataset of densely annotated sequences, which can be used for binary classification, 
and other temporal-related tasks. 
In ($ii$) \emph{incremental} mode (which could informally be classified as ``learning over time''), LTLZinc outputs a single sequence of datasets, to be used in incremental/continual \cite{wang2024comprehensive} and curriculum learning settings. 
As LTL is strictly more expressive than current definitions of incremental learning in the literature, it is straight-forward to encode class-incremental, task-incremental and domain adaptation settings with LTLZinc.

Internally, LTLZinc converts the temporal specification $\mathcal{F}$ into a DFA annotated with constraints, and solves a collection of  MiniZinc programs, corresponding to the subset of $\mathcal{C}$ satisfying each of the transition guards, caching every solution found. Then, (positive/negative) sequences of user-specified length are generated by random walks along the automaton, where at each transition one solution of the corresponding problem is randomly sampled. Annotations contain the background knowledge, consisting of the user specification (the quadruple $\langle\mathcal{X}, \mathcal{Y}, \mathcal{C}, \mathcal{F}\rangle$) and the generated automaton, sequence-level binary labels, the trace of traversed states during generation, the trace of constraint truth values, 
and the trace of image labels for each stimulus of the $N$-uple). 

\begin{table*}
\begin{minipage}{0.27\textwidth}
\includegraphics[width=0.9\linewidth]{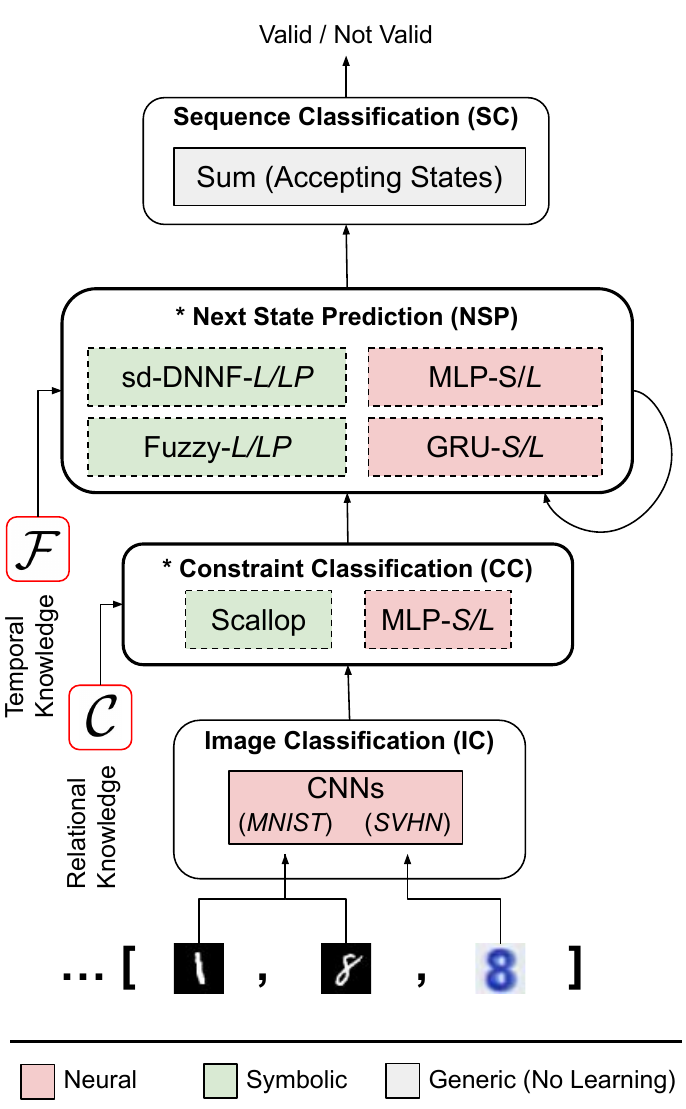}
\captionof{figure}{Stages of our architecture. We implemented {\sc cc} and {\sc nsp} in multiple ways, that we listed by means of the green (symbolic) and reddish (neural) dashed-borders sub-blocks.}
\label{fig:pipeline}
\end{minipage}
\hfill
\begin{minipage}{0.7\textwidth}
	\centering
	\resizebox{1.0\textwidth}{!}{
		\begin{tabular}{c|ll|l|llll}
			\toprule
			Task & Config. {\sc cc}-{\sc nsp} & Best Model {\sc cc} {\sc nsp} (epoch) & Avg Acc. $\uparrow$ & \textsc{ic} acc. $\uparrow$ & \textsc{cc} Acc. $\uparrow$ & \textsc{nsp} Acc. $\uparrow$ & \textsc{sc} Acc. $\uparrow$\\
			\midrule
			Task 1 & Neural-Neural & MLP-L MLP-L (24) & $0.84$ & $\textbf{0.88}$ & $0.85 \pm 0.01$ & $0.72$ & $0.90$\\
			Task 1 & Neural-Symbolic & MLP-L sd-DNNF-LP (24)* & $0.85$ & $0.86$ & $0.85$ & $0.81$ & $0.90$\\
			Task 1 & Symbolic-Neural & Scallop MLP-L (16)$^\dag$ & $0.82 \pm 0.07$ & $0.88$ & $0.91$ & $0.59 \pm 0.26$ & $0.90$\\
			Task 1 & Symbolic-Symbolic & Scallop Fuzzy-P (24)$^\dag$* & $\textbf{0.88}$ & $0.87$ & $0.91 \pm 0.01$ & $\textbf{0.84} \pm 0.01$ & $\textbf{0.91} \pm 0.01$\\
			\midrule
			Task 2 & Neural-Neural & MLP-S GRU-L (23)$^\dag$ & $0.69 \pm 0.03$ & $0.88 \pm 0.01$ & $0.87 \pm 0.01$ & $0.50 \pm 0.13$ & $0.50$\\
			Task 2 & Neural-Symbolic & MLP-S Fuzzy-LP (24)$^\dag$ & $0.71$ & $0.81 \pm 0.03$ & $0.82 \pm 0.02$ & $0.56 \pm 0.02$ & $0.67 \pm 0.03$\\
			Task 2 & Symbolic-Neural & Scallop GRU-L (22)* & $0.72 \pm 0.03$ & $0.90$ & $\textbf{0.93}$ & $0.56 \pm 0.12$ & $0.50$\\
			Task 2 & Symbolic-Symbolic & Scallop Fuzzy-P (16)$^\dag$* & $\textbf{0.79} \pm 0.01$ & $\textbf{0.89} \pm 0.01$ & $0.91$ & $\textbf{0.70} \pm 0.01$ & $\textbf{0.67} \pm 0.04$\\
			\midrule
			Task 3 & Neural-Neural & MLP-L GRU-L (24) & $0.68$ & $0.96$ & $0.82 \pm 0.01$ & $0.44$ & $0.50$\\
			Task 3 & Neural-Symbolic & MLP-L sd-DNNF-P (24)* & $0.66 \pm 0.01$ & $0.93$ & $0.71 \pm 0.02$ & $\textbf{0.51} \pm 0.03$ & $0.49 \pm 0.01$\\
			Task 3 & Symbolic-Neural & Scallop GRU-L (19)$^\dag$ & $\textbf{0.73} \pm 0.01$ & $\textbf{0.98}$ & $\textbf{0.98} \pm 0.01$ & $0.45 \pm 0.03$ & $0.50$\\
			Task 3 & Symbolic-Symbolic & Scallop sd-DNNF-LP (20)* & $0.55 \pm 0.36$ & $0.43 \pm 0.47$ & $0.64 \pm 0.28$ & $0.49 \pm 0.46$ & $\textbf{0.63} \pm 0.28$\\
 Task 3 & (Symbolic-Symbolic) & Scallop sd-DNNF-LP (20)* & ${0.93}$ & $0.96$ & $0.96$ & ${0.90}$ & ${0.91}$\\
			\midrule
			Task 4 & Neural-Neural & MLP-L GRU-S (21)$^\dag$* & $0.62 \pm 0.06$ & $\textbf{0.89} \pm 0.01$ & $0.80 \pm 0.01$ & $0.30 \pm 0.26$ & $0.50$\\
			Task 4 & Neural-Symbolic & MLP-S sd-DNNF-P (19)$^\dag$ & $0.60 \pm 0.06$ & $0.80$ & $0.62 \pm 0.04$ & $0.49 \pm 0.11$ & $0.48 \pm 0.12$\\
			Task 4 & Symbolic-Neural & Scallop GRU-L (20)$^\dag$* & $0.68 \pm 0.01$ & $0.88 \pm 0.01$ & $\textbf{0.86} \pm 0.02$ & $0.46 \pm 0.02$ & $0.50$\\
			Task 4 & Symbolic-Symbolic & Scallop sd-DNNF-P (16)$^\dag$* & $\textbf{0.74} \pm 0.05$ & $0.83 \pm 0.01$ & $0.79 \pm 0.01$ & $\textbf{0.69} \pm 0.08$ & $\textbf{0.67} \pm 0.09$\\
			\midrule
			Task 5 & Neural-Neural & MLP-L MLP-L (24)* & $0.61 \pm 0.01$ & $0.95 \pm 0.01$ & $0.57 \pm 0.04$ & $0.42 \pm 0.03$ & $0.50$\\
			Task 5 & Neural-Symbolic & MLP-L sd-DNNF-P (4) & $0.54 \pm 0.01$ & $0.94 \pm 0.01$ & $0.50 \pm 0.05$ & $0.23$ & $0.50$\\
			Task 5 & Symbolic-Neural & Scallop MLP-L (24)$^\dag$* & $\textbf{0.88} \pm 0.06$ & $\textbf{0.98}$ & $\textbf{0.96}$ & $\textbf{0.78} \pm 0.02$ & $\textbf{0.80} \pm 0.26$\\
			Task 5 & Symbolic-Symbolic & Scallop Fuzzy-P (23) & $0.85 \pm 0.01$ & $0.98$ & $0.96$ & $0.74 \pm 0.02$ & $0.71 \pm 0.05$\\
			\midrule
			Task 6 & Neural-Neural & MLP-L MLP-L (22) & $0.67 \pm 0.01$ & $0.96$ & $0.74 \pm 0.01$ & $0.48 \pm 0.10$ & $0.48 \pm 0.06$\\
			Task 6 & Neural-Symbolic & MLP-S Fuzzy-P (22) & $0.56 \pm 0.09$ & $0.81 \pm 0.15$ & $0.53 \pm 0.12$ & $0.39 \pm 0.06$ & $0.52 \pm 0.03$\\
			Task 6 & Symbolic-Neural & Scallop GRU-L (16)$^\dag$ & $0.80 \pm 0.07$ & $\textbf{0.98}$ & $\textbf{0.98} \pm 0.01$ & $0.63 \pm 0.18$ & $0.62 \pm 0.11$\\
			Task 6 & Symbolic-Symbolic & Scallop sd-DNNF-P (21)$^\dag$* & $\textbf{0.85} \pm 0.05$ & $0.97$ & $0.97$ & $\textbf{0.70} \pm 0.08$ & $\textbf{0.76} \pm 0.10$\\
			
			\bottomrule
		\end{tabular}
	}
	\caption{\textbf{Q2}. Test set accuracies (mean $\pm$ std--if non-zero--, 3 runs, with a sample run in brackets) of the pipeline in Fig.~\ref{fig:pipeline} (Avg Acc.) and its stages, for different of {\sc cc}'s and {\sc nsp}'s (i.e., Neural-Symbolic stands for Neural {\sc cc} and Symbolic {\sc nsp}). Markers: $^{*}$ = semantic loss; $^{\dag}$ = calibrated. Best model (named ``{\sc cc} model {\sc nsp} model'')  and epoch selected by Avg Acc on a validation set.}
 \label{tab:main}
\end{minipage}
\end{table*}

\section{Methodology}
\label{sec:setup}

We model the whole sequence classification task by 
%
a multi-stage pipeline composed of the following sub-tasks, also sketched in Fig.~\ref{fig:pipeline}: ({\sc\small ic}) image classification, mapping data from each $\mathcal{X}_i$ to the corresponding $\mathcal{Y}_i$; ({\sc\small cc}) constraint classification, leveraging ``relational'' knowledge; ({\sc\small nsp}) next state prediction, leveraging ``temporal'' knowledge; ({\sc\small sc}) sequence classification, i.e., the binary classification problem.
LTLZinc provides annotations for each of these sub-tasks, so that we can evaluate performance both in isolation for each component, and jointly in an end-to-end fashion
, while also being able to define multiple training objectives, applied at different stages.
%
We 
address the following research questions.

\noindent\fbox{%
    \parbox{0.96\columnwidth}{%
        \small
        \textbf{Q1: Can neural-only methods solve the proposed tasks?} Given enough training capacity and trainable parameters, can a neural network successfully perform sequence classification, without exploiting background knowledge?\\
        \textbf{Q2: How do neural-only and neuro-symbolic methods compare?} Can background knowledge overcome the issues caused by error amplification along our multi-stage architecture? Are neuro-symbolic methods more prone to training instability than neural-only approaches? \\
        \textbf{Q3: What is the effect of upstream noise on exact symbolic methods?} Do wrong, but confident, predictions of symbols $y_i$'s affect downstream accuracy more than correct predictions associated with lower confidence? How is noise propagated? 
    }%
}

\noindent We create the context to address these questions by detailing the stages of the computational pipeline we implemented.

\parafango{{\sc (ic)} Image classification} The first stage corresponds to a traditional neural-network-based image classification task, estimating the probability of the symbol/class assignment $[y_{0}^{t}, \ldots, y_{N-1}^{t}]$, i.e., $P_{\text{\sc\tiny ic}}([y_{0}^{t}, \ldots, y_{N-1}^{t}] \mid [x_{0}^{t}, \ldots, x_{N-1}^{t}])$. 
Since the proposed LTLZinc tasks (Appendix \ref{app:tasks}) consist of perceptually simple images, our implementation is based on a popular convolutional architecture (details in Appendix~\ref{app:backbone}), for each perceptual domain 
(e.g., for the example of Section~\ref{sec:seq}, we have two instances of the same architecture: one to predict symbols $y_A, y_B$ (MNIST categories) and another one for symbols $y_C$ (SVHN categories), see Fig.~\ref{fig:pipeline}).

\parafango{{\sc (cc)} Constraint classification} The set of all the image classes  
predicted by the \textsc{ic} module, is mapped to $|\mathcal{C}|$ validity values, 
each of them indicated with $\beta_i$, one for each of the constraints defined in the relational knowledge $\mathcal{C}$. This stage estimates $P_{\text{\sc\tiny cc}}([\beta_{0}^{t}, \ldots, \beta_{|\mathcal{C}| - 1}^{t}] \mid [y_{0}^{t}, \ldots, y_{N-1}^{t}])$. As 
relational knowledge is
explicitly provided in our setting, this map 
can be implemented with any knowledge injection technique.
Our experiments focus on tasks characterized by constraints which can be expressed in Datalog.
Hence, we chose Scallop~\cite{li2023scallop}, a neuro-symbolic engine capable of probabilistic reasoning over Datalog programs, supporting inference over multiple provenance semirings~\cite{green2007provenance}.\footnote{In initial explorations, the default top-$k$ proofs provenance ($k=1$) resulted in the best trade-off between inference time and accuracy.} 
Scallop programs are differentiable end-to-end, however they do not possess trainable parameters. To increase module flexibility, we augment our architecture with an optional set of additional learnable calibration parameters 
(enabled by a ``calibrate'' hyper-parameter in the experiments of Section \ref{sec:exp}), which independently apply temperature rescaling to both input and output probabilities. 
Learning takes place jointly with the \textsc{ic} module, and it allows us to control both the entropy of the distributions and the confidence of prediction, while preserving the argmax.
%
We compare Scallop with two fully-learning-based approaches: small multi-layer perceptron with limited capacity (8 hidden neurons, MLP-S) and a larger one (64 hidden neurons, MLP-L), see Appendix \ref{app:constraints}.

\parafango{{\sc (nsp)} Next state prediction} The temporal reasoning component is rooted on the definition of a discrete space of $M$ elements, each consisting of a state of the DFA-equivalent to the temporal specification $\mathcal{F}$, based on the observed validity values of relational constraints, $\beta_i^t$'s.
In this way, the temporal reasoning task boils down to a next-state prediction problem, i.e., a recurrent classification problem where $\alpha^t \in [0, M-1]$ is the predicted class/next-state, with probability $P_{\text{\sc\tiny nsp}}(\alpha^{t} \mid [\beta_{0}^{t}, \ldots, \beta_{\mathcal{V}-1}^{t}], \alpha^{t-1})$.\footnote{Here and in the following, $t = -1$ is the initial instant, where the automaton of the \textsc{nsp} module is initialized to state ``$0$''.} 
We compare four learning-based approaches (two multi-layer perceptrons, namely MLP-S and MLP-L, and two gated recurrent units, GRU-S, GRU-L, with 8 and 64 hidden/state neurons, respectively), and four symbolic approaches.
GRUs are augmented with a simple encoder-decoder, to convert the continuous hidden state into $M$ discrete classes.
For the symbolic approaches, we encode the ground truth finite state automaton, following the works of~\cite{umili2023grounding} and~\cite{manginas2024nesya}. 
Such automaton is a set of propositional logic formulas as $\textsc{\small prev\_state} \wedge \textsc{\small trans\_label} \rightarrow \textsc{\small next\_state}$, encoded by logic tensor networks~\cite{badreddine2022logic} (here named Fuzzy) and sd-DNNF~\cite{darwiche2002knowledge}, respectively. To assess the effect of numerical stability, we performed computations both in probability space (conjunction as product, disjunction as sum) and log-probability space (conjunction as sum, disjunction as logsumexp), adding the suffix -P and -LP to distinguish the two cases, respectively.
As in the \textsc{cc} module, when ``calibrate'' is set to true, automata-based \textsc{nsp}'s predictions are temperature-calibrated.

\parafango{{\sc (sc)} Sequence classification} The final output $\alpha^{T-1}$ of {\sc nsp} directly encodes the final state of the automaton. Hence, we can perform sequence classification in closed form, by computing $P_{\text{\sc\tiny sc}}(f(\mathcal{S}) = 1 \mid \alpha^{T-1}) = \sum\limits_{s \in \text{\sc\tiny accepting}} P(\alpha^{T-1} = s)$. 

\parafango{Training the pipeline}
The four stages are combined into: 
\begin{equation}
\nonumber
    P(f(\mathcal{S}) = 1) =  \left( \prod_{t=0}^{T-1} P^t_{\text{\sc\tiny ic}} P^t_{\text{\sc\tiny cc}} P^t_{\text{\sc\tiny nsp}} \right) P_{\text{sc}}(f(\mathcal{S}) = 1 \mid \alpha^{T-1}),
\end{equation}
where $P^t_{\cdot}$ is the shorthand notation for the already introduced probabilities with arguments at time $t$.
We train our multi-stage architectures with four loss functions, each weighted by a $\lambda_{\cdot}$ hyper-parameter. \textsc{ic} and \textsc{nsp} (the latter conditioned on the previous state) exploit a 
categorical cross-entropy loss, 
while \textsc{cc} and \textsc{sc} 
a binary cross-entropy loss. 
For \textsc{sc}, we also evaluate the semantic loss proposed by~\cite{umili2023grounding}.
Preliminary experiments demonstrated that training diverges due to extremely low initial confidence in image classification,\footnote{This is especially true when the \textsc{cc} module is a Scallop program, as at the beginning of training, misclassified images lead to confidently predicting violated constraints most of the time.} 
making the optimizer unable to converge, except in very simple tasks. This behavior is well known in the literature~\cite{manhaeve2021approximate,van2024independence,maene2024hardness}, and, in this paper, it only affects experiments aimed at investigating \textbf{Q2}, where we bootstrap the  \textsc{ic} module with a $5$ epochs pre-training phase, using the \textsc{ic}-loss only, 
to ensure a good starting image classification. 

\section{Experiments}\label{sec:exp}
\begin{figure*}
	\centering
    \begin{minipage}{0.65\linewidth}
    	\includegraphics[width=1.0\textwidth]{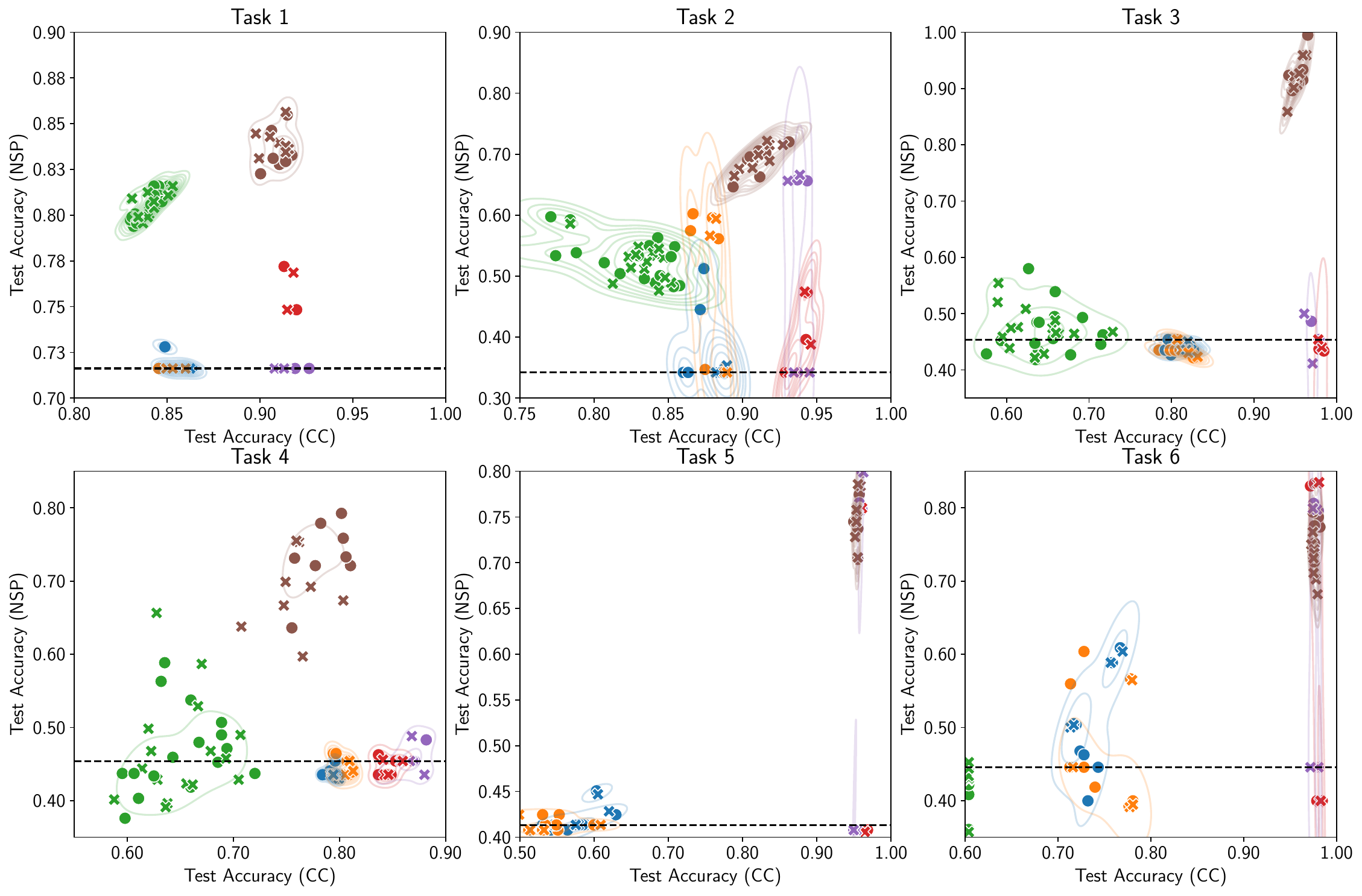}
           \end{minipage}
 \begin{minipage}{0.14\linewidth}
    \raisebox{0.5\height}{\includegraphics[width=1.0\textwidth,trim={0, 1.5cm, 0, 0},clip]{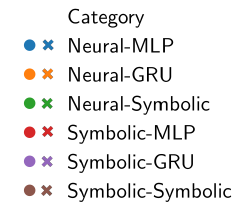}}\\
{\includegraphics[width=1.0\textwidth,trim={0, 0, 0, 3.5cm},clip]{imgs/const-vs-succ_legend.pdf}}    
    \end{minipage}
    \vskip -3mm
	\caption{\textbf{Q2}. {\sc cc}-{\sc nsp} accuracy trade-off for different families of architectures (i.e., Neural/Symbolic {\sc cc}, Neural/Symbolic {\sc nsp}, where Neural {\sc nsp} is indicated either by MLP or GRU, to compare them as well). The horizontal dashed line indicates the performance of a deterministic baseline of {\sc nsp} always choosing the successor state most represented in the training set.} 
	\label{fig:main}
\end{figure*}
We hereby present an experimental evaluation conducted on six LTLZinc tasks.\footnote{LTLZinc generator, our dataset, and code for full reproducibility: \url{https://github.com/continual-nesy/LTLZinc}.}
Each of the six tasks consists in a binary classification problem on $400$ image sequences of variable length (between 10 and 20), where
images are sampled from the MNIST~\cite{lecun1998mnist} and the Fashion MNIST~\cite{xiao2017fmnist} datasets.
Tasks are annotated with relational and temporal knowledge (see Appendix~\ref{app:tasks} for all the details).
We address research questions {\bf Q1}, {\bf Q2}, {\bf Q3} of Section~\ref{sec:setup} with three specific experimental activities, aimed to study the effect of different architectures and available knowledge. Notice that our goal is not to simply find the best performing model, but to explore the different scenarios and gain insights on our challenging setting.
In Appendices~\ref{app:main}, \ref{app:neur} and~\ref{app:abla} we report values of all involved hyper-parameters.

\parafango{Q1}
To explore neural-only methods, our modular architecture is instantiated with neural components only (i.e., the convolutional backbone (\textsc{ic}), MLP-S/L for constraint prediction (\textsc{cc}), and MLP-S/L or GRU-S/L for next-state prediction (\textsc{nsp}), Fig.~\ref{fig:pipeline}). Supervision is provided at every level, weighting each loss by the same positive coefficient ($\lambda_{\cdot} = 1.0$). 
Our expectations were that the neural-only pipeline would have been good in managing the \textsc{ic} stage, but it was not clear how it would have performed in the other ones. Results (detailed in Appendix~\ref{app:neuronly}) confirmed that, despite their perceptual simplicity, no task can be effectively learned with sequence-level labels alone. 
We experienced slow convergence and sub-optimal performance at the end of training. In fact, in spite of good overall image classification, the optimizer is often stuck in local equilibrium points, causing performance for other training objectives to plateau multiple times during training. Harder reasoning tasks exacerbate this effect. 

\parafango{Q2}
This second batch of experiments aims to compare neural vs. neuro-symbolic methods, thus the effect of background knowledge over multiple reasoning steps, as well as the interaction between neural and symbolic components.
We fix a neural perceptual backbone (\textsc{ic}), and then build four model categories by combining either Neural (red, Fig.~\ref{fig:pipeline}) or Symbolic (green, Fig.~\ref{fig:pipeline}) modules for constraint prediction (\textsc{cc}) and temporal reasoning (\textsc{nsp}). 
We will indicate henceforth each configuration with a shorthand notation, e.g., Symbolic-Neural means Symbolic \textsc{cc} and Neural \textsc{nsp} modules.
The use of the semantic loss and/or of calibration (Section~\ref{sec:setup}), are treated as additional hyper-parameters.
Training is performed in two steps: 5 epochs of pre-training for the perceptual module, and then 20 epochs of training for the entire architecture. Overall, these experiments confront 96 combinations, across 6 tasks, each seeded 3 times.

\parafango{Q3}
The third batch of experiments focuses on symbolic-only methods, thus on the Symbolic-Symbolic (green, Fig.~\ref{fig:pipeline}) architectures. We exploit probability calibration (Section~\ref{sec:setup}) as the only form of learning, to minimize noise and precisely pinpoint the effect of variables such as upstream label uncertainty, and supervision ``distance'' from the reasoning component. We also consider replacing {\sc ic} and {\sc cc} modules with specific ``oracles'' returning the ground truth labels, possibly with some level of corruption. In particular, we either replace only \textsc{ic} or both \textsc{ic} and \textsc{cc}. 
An oracle is characterized by two hyper-parameters: how ground truth is corrupted before feeding the next module, and the amount of corruption in terms of noise probability $p$. ``Flip oracles'' return correct labels with probability $1-p$, and random labels with probability $p$, both with confidence $1.0$. ``Confidence oracles'', on the other hand, always return correct labels with random confidence between $1-p$ and $1.0$, redistributing the remaining mass to other labels.
When $p = 0.0$, both oracles yield ground truth labels with maximum confidence, referred to as ``perfect oracle''.
%
\begin{figure*}
	\centering
	\includegraphics[width=.216\linewidth,trim={0, 0, 28cm, 0},clip]{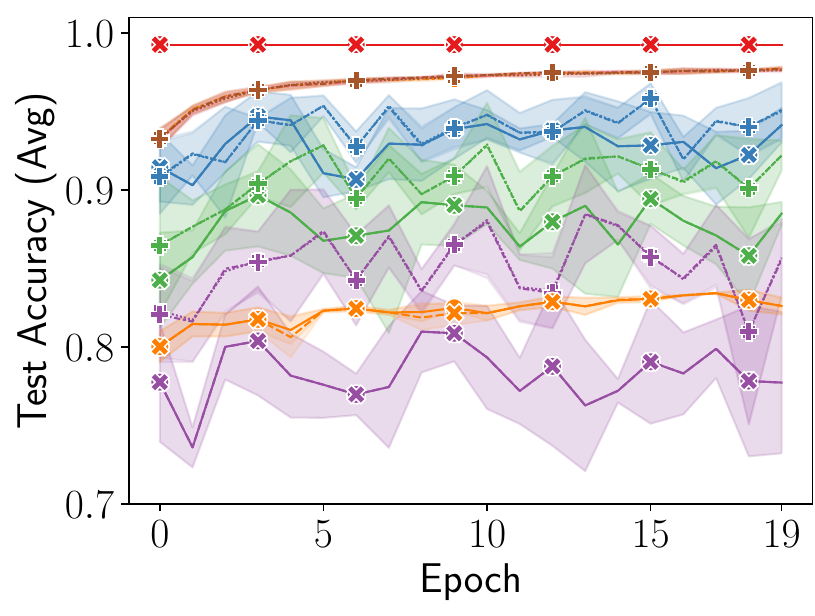}
    \includegraphics[width=.4356\linewidth,trim={13.8cm, 0, 0, 0},clip]{imgs/ablation_task4.pdf}
    \hfill
    \raisebox{0.5cm}{\includegraphics[width=.3\linewidth,trim={0, 0, 0, 0},clip]{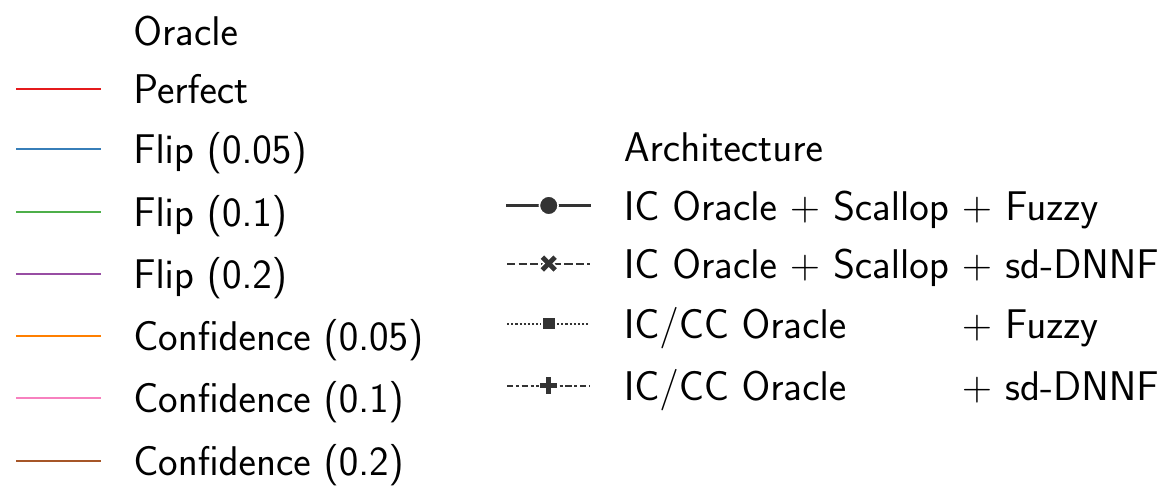}}
    \vskip -2mm
	\caption{\textbf{Q3}. Accuracies for \textit{Task 4} with oracular predictors. Oracle types (perfect, flip, confidence) are described in Section~\ref{sec:exp}.}
	\label{fig:abl-task4}
\end{figure*}
\subsection{The impact of neuro-symbolic approaches}
Fig.~\ref{fig:main} highlights the trade-offs between \textsc{cc} and \textsc{nsp}, as a function of the different Neural/Symbolic implementations.
Exploiting a symbolic component both for \textsc{cc} and \textsc{nsp} (Symbolic-Symbolic, brown points) allows to achieve the best trade-off for every task considered. Unlike other combinations, these approaches consistently outperform naive baselines which always return the most probable class observed in the training set, both for \textsc{nsp} (dashed horizontal line), and \textsc{cc} (not shown, outside the left boundary of plots).
The additional learning capacity provided by temperature calibration (dots vs. crosses) has an overall limited effect.
With the exception of tasks 1, 2 and 4, neural modules for \textsc{nsp} (clusters Symbolic-MLP/GRU and Neural-MLP/GRU, blue, orange, red and purple) achieve unsatisfactory performance, even when fed from highly-accurate symbolic \textsc{cc} predictions. These architectures, at times, perform on par with the most probable guessing baseline, and often below it.
Symbolic modules for \textsc{nsp} are characterized by a large dispersion, due to optimization challenges: this is especially true when exploiting neural \textsc{cc} modules (Neural-Symbolic, green), but it can also be observed in combination with symbolic \textsc{cc} modules (Symbolic-Symbolic, brown). Conversely, neural \textsc{nsp}, even though often showing unsatisfactory performance, is characterized by a much smaller inter-experiment variance.
%

%
Table~\ref{tab:main} summarizes performance for each task (each group shows the best set of hyper-parameters). 
Overall, the Symbolic-Symbolic category dominates over other groups, with the only exception of task 5, where it performs slightly worse than the Symbolic-Neural family. When observing constraint accuracy alone, a downstream Symbolic \textsc{nsp} module often negatively affects performance, with task 4 presenting the highest delta between Symbolic-Symbolic and Symbolic-Neural.
Conversely, when focusing on next state prediction, it is clear that an upstream Symbolic \textsc{cc} module is beneficial to a downstream Symbolic temporal reasoning module, however this clear-cut performance improvement is flipped when the \textsc{nsp} module is Neural. This behavior hints at complex interactions between the constraint and temporal modules, where architectural choices bi-directionally affect both components.
No clear trend in the effect of other hyper-parameters can be identified. Early stopping is almost always triggered near the end of training, hinting at a possibly incomplete convergence. Temperature calibration is often selected, but performance improvements are minor (cfr. with Fig. \ref{fig:main}), and there is no proof that using the semantic loss proposed by~\cite{umili2023grounding} is significantly more beneficial than traditional binary cross-entropy in our setting.
Variances across three runs for each set of hyper-parameters are low. Comparing this observation to Fig.~\ref{fig:main} (which, instead, highlights hyper-parameter-based variance), it can be observed that convergence in multi-stage neuro-symbolic sequence classification is heavily dependent on architectural choices, but relatively unaffected by parameter initialization.
Task 3 is an exception and it presents a failure case of the Symbolic-Symbolic family, due to training instability. When able to converge, however, it is the only approach capable of improving over random guessing performance for sequence classification for this task (we reported this sample result surrounding the model name with brackets).

\subsection{The impact of oracular predictors}

Fig.~\ref{fig:abl-task4} shows the effect on an exemplar task (Task 4) of oracular predictors on symbolic temporal reasoning modules.
Figures \ref{fig:abl-avg}, \ref{fig:abl-const} and \ref{fig:abl-succ}, in the supplementary materials, show the same plots for every task.
In general, the trainable temperature parameters are quickly optimized, with curves following a mostly horizontal trend. However, flip oracles (blue, green and purple lines) are characterized both by larger variance and inter-epoch oscillations, compared to confidence oracles (orange, pink and brown), which present a remarkably stable behavior. This effect is present in every task, regardless of reasoning difficulty.
Noise injection, however, affects the two oracles differently: performance for flip oracles degrades linearly when increasing noise, while confidence oracles are characterized by non-linear behavior. Small amounts of noise do not affect confidence oracles appreciably, while larger amounts tend to harm performance more than the flip oracle. This effect is more evident for harder reasoning tasks: for instance, in task 4, a noise of $p = 0.1$ is enough to cause random guessing performance (Fig. \ref{fig:abl-task4-nolimits} in the Appendix).
%
With oracles, the effect of different automata encodings is virtually non-existent (fuzzy and sd-DNNF markers overlap almost everywhere). 
\textsc{IC} oracles are consistently more affected by noise, compared to \textsc{IC/CC} ones. This behavior seems counterintuitive, as the Scallop module performs near-exact inference; however, it is in line with the hypothesis of uncertainty accumulation over multiple steps of reasoning.

When focusing on constraint performance (Fig.~\ref{fig:abl-const} in the supplementary materials), flip oracles tend to have stable performance, while confidence oracles can exploit temperature parameters to achieve a small learning capacity (slight upward trend across epochs).
In general, flip oracles are more robust to noise, with confidence oracles achieving unsatisfactory performance (outside plot boundaries, see Fig. \ref{fig:abl-task4-nolimits} in the Appendix) for harder tasks or higher ($p > 0.05$) degrees of uncertainty.
This behavior clashes with the desirable property of predictive confidence correlating with uncertainty, which is one of the advantages of neuro-symbolic AI, compared to (uncalibrated) neural networks. Flip oracles achieving better performance than confidence oracles, for similar levels of noise, hint at the fact that an overconfidently-wrong classifier can be more successful than a reluctantly-correct one, in temporal reasoning settings.
%
%

\section{Conclusions}
We extended knowledge-driven sequence classification to a relational setting, introducing a novel benchmarking framework and baselines for neural-only and multi-stage neuro-symbolic methods in a knowledge-driven, and long-temporal-horizon regime.
Experiments outline a challenging setting, where neural networks struggle to generalize and state-of-the-art neuro-symbolic methods fall short when stacked: temporal reasoners extended to a relational domain can fail even with full background knowledge available, and general-purpose first-order reasoners suffer from training instabilities in recurrent settings.
We argue that the proposed benchmarking framework can benefit the neuro-symbolic and temporal reasoning communities, further pushing research boundaries towards more expressive frameworks, tighter neuro-symbolic integration, and more robust time-driven approaches.

\appendix
\section*{Acknowledgements}
%
M.L. was supported by CAI4DSA actions (Collaborative Explainable neuro-symbolic AI for Decision Support Assistant), PARTENARIATO ESTESO ``Future Artificial Intelligence Research - FAIR'', SPOKE 1 ``Human-Centered AI'' Universit\`a di Pisa, CUP B13C23005640006. The scholarship by L.S.L. was funded by the Italian Ministry of University and Research (DM 351/2022, PNRR).
S.M. was supported by the University of Siena (Piano per lo Sviluppo della Ricerca - PSR 2024, F-NEW FRONTIERS 2024), under the project ``TIme-driveN StatEful Lifelong Learning'' (TINSELL) and also by the project ``CONSTR: a COllectionless-based Neuro-Symbolic Theory for learning and Reasoning'', PARTENARIATO ESTESO ``Future Artificial Intelligence Research - FAIR'', SPOKE 1 ``Human-Centered AI'' Universit\`a di Pisa,  ``NextGenerationEU'', CUP I53C22001380006.
L.S.L. scholarship was funded by the Italian Ministry of University and Research (DM 351/2022, PNRR).
We kindly thank Luc De Raedt and Nikolaos Manginas for their insightful comments.

\bibliographystyle{named}
\bibliography{biblio}

\end{document}


\maketitle
\appendix

\section{Task Specifications}\label{app:tasks}
\let\originalthefigure\thefigure
\let\originalfigurename\figurename
\setcounter{figure}{0}
\renewcommand{\figurename}{Task}
We experiment on six sequence classification tasks where positive sequences satisfy the temporal specification, while negative sequences violate it. Note that this means that negative sequences may still satisfy the provided constraints, albeit at the ``wrong'' time.

For each task, we generate 400 sequences (320 train, 40 val, 40 test) of variable length, between 10 and 20 time steps.
Images are sampled from the MNIST digits~\cite{lecun1998mnist} and the Fashion MNIST~\cite{xiao2017fmnist} datasets.
Images for each split are sampled according to the original image classification dataset splits (i.e., for MNIST and FMNIST datasets,  we sample training and validation set from the 60,000 training images and the test set from the 10,000 test images).
Each sequence is associated with the following annotations: image labels for each time step, constraint validity for each time step, automaton state trace, sequence label. The automaton used to generate each task is also available as background knowledge.

In the following, we formally describe the 6 task we considered in our experimentation.
With an abuse of notation, we condense perceptual  and symbolic domains $\mathcal{X}, \mathcal{Y}$ by indicating only the former (i.e., $Z =  \mathimg{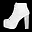}$ corresponds to $\langle \mathcal{X}_Z = \mathimg{fmnist1}, \mathcal{Y}_Z = \text{``boot''}  \rangle$). Symbolic classes  are enumerated in lexicographic order (i.e., ``bag'' = 0, ``boot'' = 1, etc.).

\begin{figure}[H]
\centering
{\bf TASK 1}
 \begin{align*}
 &\ltlglobally (p \leftrightarrow \ltlnext\ltlnext q)\\
 \\
 p := &Y < Z\\
 q := &\texttt{all\_equal}([V,W,X])\\
 Y, Z \in& \{\mathimg{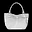}, \mathimg{fmnist1}, \mathimg{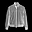}, \mathimg{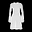}, \mathimg{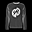}, \mathimg{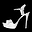}, \mathimg{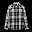}, \mathimg{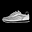}, \mathimg{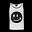}, \mathimg{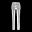}\}\\
 V, W, X \in& \{\mathimg{fmnist5}, \mathimg{fmnist6}, \mathimg{fmnist7}, \mathimg{fmnist8}, \mathimg{fmnist9}\}\\
 \end{align*}
 \vskip -0.7cm
 \caption{It will always be the case that $Y$ is less than $Z$ if and only if in two steps ahead, $V$, $W$ and $X$ all belong to the same class. The corresponding automaton has 8 states.}
\end{figure}

\begin{figure}[H]
\centering
{\bf TASK 2}
 \begin{align*}
 &\ltlglobally ((p \wedge \ltlnext p \wedge \ltlnext \ltlnext p) \rightarrow\ltlnext\ltlnext\ltlnext q)\\
 \\
 p := &Y < Z\\
 q := &\texttt{all\_equal}([V,W,X])\\
 Y, Z \in& \{\mathimg{fmnist0}, \mathimg{fmnist1}, \mathimg{fmnist2}, \mathimg{fmnist3}, \mathimg{fmnist4}, \mathimg{fmnist5}, \mathimg{fmnist6}, \mathimg{fmnist7}, \mathimg{fmnist8}, \mathimg{fmnist9}\}\\
 V, W, X \in& \{\mathimg{fmnist5}, \mathimg{fmnist6}, \mathimg{fmnist7}, \mathimg{fmnist8}, \mathimg{fmnist9}\}\\
 \end{align*}
  \vskip -0.7cm
 \caption{It will always be the case that if $Y$ is less than $Z$ for three consecutive steps, then in the step immediately ahead, $V$, $W$ and $X$ all belong to the same class. The corresponding automaton has 5 states.}
\end{figure}

\begin{figure}[H]
\centering
{\bf TASK 3}
 \begin{align*}
 &\ltlfinally p \wedge (q \ltluntil \ltlnext p)\\
 \\
 p := &\texttt{all\_different}([X,Y,Z])\\
 q := &X < Y + Z\\
 X, Y, Z \in& \{\mathimg{mnist0}, \mathimg{mnist1}, \mathimg{mnist2}, \mathimg{mnist3}, \mathimg{mnist4}, \mathimg{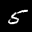}, \mathimg{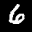}, \mathimg{mnist7}, \mathimg{mnist8}, \mathimg{mnist9}\}\\
 \end{align*}
  \vskip -0.7cm
 \caption{It will hold that $X < Y + Z$ up to the timestep immediately before the point in which $X$, $Y$, and $Z$ all belong to different classes. The latter condition is guaranteed to be verified at somepoint in the sequence. The corresponding automaton has 5 states.}
\end{figure}

\begin{figure}[H]
\centering
{\bf TASK 4}
 \begin{align*}
 &\ltlfinally p \wedge (q \ltluntil \ltlnext p)\\
 \\
 p := &\texttt{all\_different}([X,Y,Z])\\
 q := &X < Y + Z\\
 X \in& \{\mathimg{mnist0}, \mathimg{mnist1}, \mathimg{mnist2}, \mathimg{mnist3}, \mathimg{mnist4}, \mathimg{mnist5}, \mathimg{mnist6}, \mathimg{mnist7}, \mathimg{mnist8}, \mathimg{mnist9}\}\\
 Y, Z \in& \{\mathimg{fmnist0}, \mathimg{fmnist1}, \mathimg{fmnist2}, \mathimg{fmnist3}, \mathimg{fmnist4}, \mathimg{fmnist5}, \mathimg{fmnist6}, \mathimg{fmnist7}, \mathimg{fmnist8}, \mathimg{fmnist9}\}\\
 \end{align*}
 \vskip -0.7cm
 \caption{It will hold that $X < Y + Z$ up to the timestep immediately before the point in which $X$, $Y$, and $Z$ all belong to different classes. The latter condition is guaranteed to be verified at some point in the sequence. This is the same as the previous task--different datasets. The corresponding automaton has 5 states.}
\end{figure}

\begin{figure}[H]
\centering
{\bf TASK 5}
 \begin{align*}
 &\ltlglobally (p \leftrightarrow \ltlweaknext \neg p)\\
 \\
 p := &W + X = Y + Z\\
 W, X, Y, Z \in& \{\mathimg{mnist0}, \mathimg{mnist1}, \mathimg{mnist2}, \mathimg{mnist3}, \mathimg{mnist4}, \mathimg{mnist5}, \mathimg{mnist6}, \mathimg{mnist7}, \mathimg{mnist8}, \mathimg{mnist9}\}\\
 \end{align*}
 \vskip -0.7cm
 \caption{The sequence alternates between satisfying $W + X = Y + Z$ and violating it. The corresponding automaton has 4 states.}
\end{figure}

\begin{figure}[H]
\centering
{\bf TASK 6}
 \begin{align*}
 &\ltlglobally (p \leftrightarrow \ltlnext q)\\
 \\
 p := &X + Y = Z\\
 q := &X + Y = 2Z\\
 X, Y, Z \in& \{\mathimg{mnist0}, \mathimg{mnist1}, \mathimg{mnist2}, \mathimg{mnist3}, \mathimg{mnist4}, \mathimg{mnist5}, \mathimg{mnist6}, \mathimg{mnist7}, \mathimg{mnist8}, \mathimg{mnist9}\}\\
 \end{align*}
 \vskip -0.7cm
 \caption{The sequence alternates between satisfying $X + Y = Z$ and $X + Y = 2Z$. The corresponding automaton has 4 states.}
\end{figure}

\setcounter{figure}{0}
\renewcommand{\figurename}{\originalfigurename}

\section{Hyper-parameters}
For simplicity, we describe each module of or pipeline (IC, CC, NSP) with a code-oriented syntax that is widely diffused in the AI community.

\subsection{\textsc{IC} Module}\label{app:backbone}
Each variable (input image) is processed by a small convolutional backbone for image classification.

\begin{align*}
 &\mathrm{Conv2d(3, 32, kernel\_size=5)}\\
 &\mathrm{MaxPool2d(kernel\_size=2)}\\
 &\mathrm{Conv2d(32, 64, kernel\_size=5)}\\
 &\mathrm{MaxPool2d(kernel\_size=2)}\\
 &\mathrm{Flatten()}\\
 &\mathrm{ReLU()}\\
 &\mathrm{Linear(64 \cdot 5 \cdot 5, 1024)}\\
 &\mathrm{Dropout(0.5)}\\
 &\mathrm{Linear(1024, num\_classes)}
\end{align*}

When variables share the same dataset, their corresponding backbones also share trainable parameters, while different datasets are processed by a different instance of the same architecture. In cases like Tasks 1 and 2, where some variables use a subset of others, we still exploit parameter sharing, suppressing the unused output neurons of the last linear layer.

\subsection{\textsc{CC} Modules}\label{app:constraints}
\subsubsection{MLP}
Each constraint is processed by the following architecture. Different constraints do not share trainable parameters.
\begin{align*}
 &\mathrm{Linear(num\_classes, num\_neurons)}\\
 &\mathrm{ReLU()}\\
 &\mathrm{Linear(num\_neurons, 1)}\\
 &\mathrm{Sigmoid()}\\
\end{align*}

If temperature calibration is enabled, the last linear layer output is divided by a separate trainable parameter, before the sigmoid layer.

\subsubsection{Scallop program}

\begin{align*}
 &\mathrm{Softmax()}\\
 &\mathrm{Scallop\_Program()}.
\end{align*}

The Scallop program is manually provided for each task.\footnote{Although every task considered in this paper can be encoded by a Scallop program, LTLZinc also allows to define constraints requiring more expressive frameworks.} As an example, the following is the program corresponding to Task 3:
\begin{verbatim}
type var_x(u8), 
 var_y(u8), 
 var_z(u8)
type p_0(), 
 p_1()

\\ All different constraint:
rel p_0() = var_x(x), 
 var_y(y), 
 var_z(z), 
 x != y, y != z, x != z

rel p_1() = var_x(x), 
 var_y(y), 
 var_z(z), 
 x < y + z
\end{verbatim}

If temperature calibration is enabled, both input logits and output probabilities are rescaled by two independent trainable parameters. Since Scallop returns probabilities, scaling is performed by first transforming the output in logit space, and then back to probabilities: $\mathrm{new\_prob} = \sigma(\frac{\sigma^{-1}(\mathrm{old\_prob})}{\mathrm{temp}})$.

Note that, if calibration is disabled, a Scallop constraint module has no learnable parameter.

\subsection{\textsc{NSP} Modules}\label{app:nextstate}
Next state prediction modules are recurrently fed with both the \textsc{CC} module output and their own past prediction.

\subsubsection{MLP}
\begin{align*}
 &\mathrm{Linear(num\_states + num\_constraints, num\_neurons)}\\
 &\mathrm{ReLU()}\\
 &\mathrm{Linear(num\_neurons, num\_states)}\\
 &\mathrm{next\_state = Softmax()}\\
 &\mathrm{sequence\_label = Sum(successor[:, accepting\_states]}\\
\end{align*}
If temperature calibration is enabled, the last linear layer output is divided by the corresponding temperature parameter, before the softmax layer.

\subsubsection{GRU}
\begin{align*}
 &\mathrm{enc = Linear(num\_states, num\_neurons)}\\
 &\mathrm{dec = Linear(num\_states, num\_neurons, num\_states)}\\
 &\mathrm{gru = GruCell(num\_constraints, num\_neurons)}\\
 \\
 &\mathrm{s0 = enc(prev\_state)}\\
 &\mathrm{s1 = gru(input, s0)}\\
 &\mathrm{next\_state = Softmax(dec(s1))}\\
 &\mathrm{sequence\_label = Sum(successor[:, accepting\_states]}\\
\end{align*}
If temperature calibration is enabled, the decoder output is divided by the corresponding temperature parameter, before the softmax layer.

\subsubsection{Fuzzy automaton}
This module follows the implementation of~\cite{umili2023grounding}. We propositionalize constraints by replacing them with unique labels, and then transpose the transition table, in order to build a separate propositional formula for each next state. Each of these formulas is the disjunction of clauses in the form: $previous\_state \wedge transition\_guard$. Finally, we use Simpy~\cite{meurer2017sympy} to simplify each formula.

During inference, we evaluate each formula by performing algebraic model counting on the probability (Fuzzy-P) or log-probability (Fuzzy-LP) semirings.

If temperature calibration is enabled, outputs are calibrated (with the probability semiring, outputs are first converted in logit space and then softmaxed, akin to the Scallop constraint module, while with the log-probability semiring, outputs are scaled directly, then log-softmaxed before being re-used as current state for the next iteration). Just like the Scallop constraint module, if temperature calibration is disabled, fuzzy automaton modules have no learnable parameters.

\subsubsection{sd-DNNF automaton}
This module follows the implementation of~\cite{manginas2024nesya}, with one major difference. In practice we start in the same way as the Fuzzy module, but instead of simplifying each formula, we compile it to a sd-DNNF (in the work of Manginas et al., only the transition guard is compiled, requiring a number of compilation steps which grows as the square of the number of states in the automaton, in our case only a linear number of compilations is needed).
Like the fuzzy modules, evaluation is performed via algebraic model counting on probability (sd-DNNF-P) and log-probability (sd-DNNF-LP) semirings.

Temperature calibration is identic to the fuzzy module.

\subsection{Main experiments (Q2) hyper-parameters}\label{app:main}
We fix the following hyper-parameters.

\noindent\textbf{Batch size.} 32

\noindent\textbf{Optimizer.} Adam

\noindent\textbf{Learning rate.} $10^{-4}$

\noindent\textbf{Pre-training learning rate.} $10^{-3}$

\noindent\textbf{Pre-training epochs.} $5$. These epochs are performed with $\lambda_\textsc{IC} = 1.0$, and every other loss disabled ($\lambda_\textsc{CC} = \lambda_\textsc{NSP} = \lambda_\textsc{SC} = 0.0$).

\noindent\textbf{Epochs.} $20$

\noindent\textbf{Gradient clipping.} $10.0$

\noindent\textbf{Image classification lambda ($\lambda_\textsc{IC}$).} $0.1$. This value is used after the end of pre-training.

\noindent\textbf{Image classification loss.} Categorical cross-entropy

\noindent\textbf{Constraint prediction lambda ($\lambda_\textsc{CC}$).} $1.0$

\noindent\textbf{Constraint loss.} Binary cross-entropy

\noindent\textbf{Next state lambda ($\lambda_\textsc{NSP}$).} $1.0$

\noindent\textbf{Next state loss.} Categorical cross-entropy

\noindent\textbf{Sequence classsification lambda ($\lambda_\textsc{SC}$).} $1.0$

\noindent\textbf{Resample images.} True. This hyper-parameter augments LTLZinc training sets by selecting at each epoch a fresh image (corresponding to the same class) from the original datasets.

The following hyper-parameters are varied across experiments. The total number of experiments performed for the main findings is 1728.

\noindent\textbf{Random seed.} 12345, 67890, 88888

\noindent\textbf{\textsc{CC} module.} MLP-S (8 neurons), MLP-L (64 neurons), Scallop (top-1 proof)

\noindent\textbf{\textsc{NSP} module.} MLP-S (8 neurons), MLP-L (64 neurons), GRU-S (8 neurons), GRU-L (64 neurons), Fuzzy-P (probability semiring), Fuzzy-LP (log-probability semiring), sd-DNNF-P (probability semiring), sd-DNNF-LP (log-probability semiring).

\noindent\textbf{Calibrate.} True, False.

\noindent\textbf{\textsc{SC} loss.} Binary cross-entropy, Fuzzy. Fuzzy loss is the one proposed by~\cite{umili2023grounding}. It is a semantic-based regularization term in product t-norm, penalizing:
\begin{align*}
 \mathcal{L}_{seq}(\mathrm{pred}, \mathrm{labels}) = &1.0 - \bigoplus \mathrm{pred_{[labels == 1]}} + \\&1.0 - \prod \mathrm{pred_{[labels == 0]}}.
\end{align*}

$\bigoplus$ is a reduction function computing the exclusive-or as: $a \oplus b = (a + b - a * b) * (1.0 - a * b)$.

\subsection{Q1 hyper-parameters}\label{app:neur}
Hyper-parameters not mentioned in this section are the same as those used in the main experiments.
We fix the following hyper-parameters.

\noindent\textbf{Epochs.} $50$

\noindent\textbf{Pre-training epochs.} $0$

\noindent\textbf{Learning rate.} $10^{-3}$

\noindent\textbf{Image classification lambda ($\lambda_\textsc{IC}$).} $1.0$

\noindent\textbf{\textsc{SC} loss.} Binary cross-entropy

\noindent\textbf{\textsc{CC} module.} MLP-L (64 neurons)

The following hyper-parameter is varied across experiments. We performed 36 experiments with purely neural baselines.

\noindent\textbf{\textsc{NSP} module.} MLP-L (64 neurons), GRU-L (64 neurons)

\subsection{Q3 hyper-parameters}\label{app:abla}
Hyper-parameters not mentioned in this section are the same as those used in the main experiments.
We fix the following hyper-parameters.

\noindent\textbf{Pre-training epochs.} $0$.

\noindent\textbf{Calibrate.} True.

The following hyper-parameters are varied across experiments, for a total of 468 experiments.

\noindent\textbf{\textsc{IC} module.} Perfect label oracle, Flip label oracle, Confidence label oracle.

\noindent\textbf{\textsc{CC} module.} Scallop (top-1 proof), Perfect constraint oracle, Flip constraint oracle, Confidence constraint oracle.

\noindent\textbf{\textsc{NSP} module.} Fuzzy-P, sd-DNNF-P.

\noindent\textbf{Oracle error.} $0.0$ (only for Perfect), $0.05$, $0.1$, $0.2$.

\section{Extended discussion about  Q1}\label{app:neuronly}

\begin{figure*}[htb!]
	\centering
	\includegraphics[width=\linewidth]{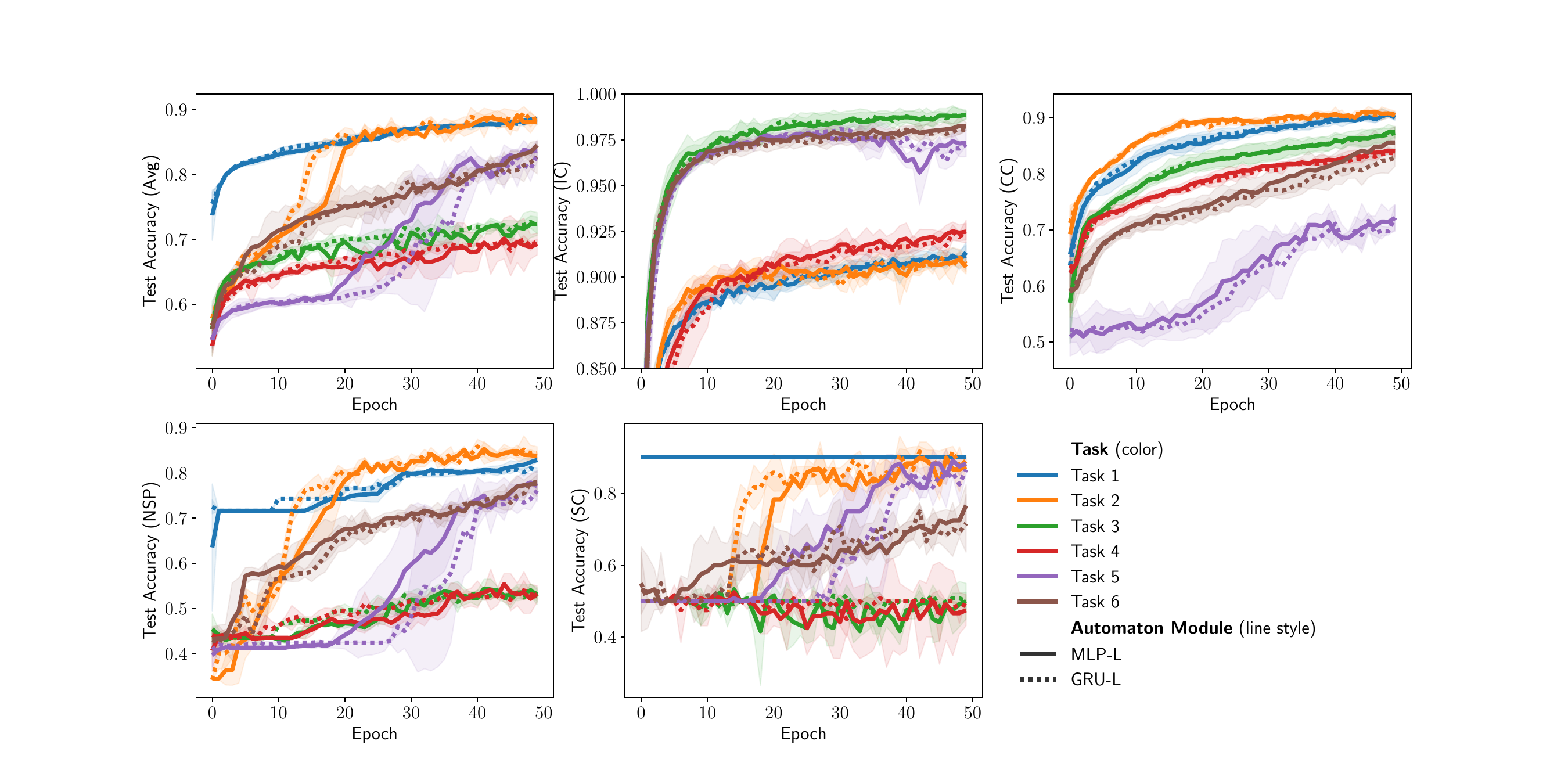}
	\caption{Neural-only experiments (\textbf{Q1}).}
	\label{fig:neural-only}
\end{figure*}

\begin{table*}[htb!]
	\centering
	\resizebox{.8\textwidth}{!}{
		\begin{tabular}{ccccccccc}
			\toprule
			Task & Automaton module & Avg acc. $\uparrow$ & Label acc. $\uparrow$ & Constraint acc. $\uparrow$ & Successor acc. $\uparrow$ & Sequence Acc. $\uparrow$ & MP Successor & MP Sequence\\
			\midrule
			Task 1 & MLP-L & $\textbf{0.89}$ & $0.91 \pm 0.01$ & $0.90 \pm 0.01$ & $\textbf{0.83} \pm 0.01$ & $0.90$ & $0.72$ & $0.90$\\
			Task 1 & GRU-L & $0.88$ & $0.91$ & $0.90 \pm 0.01$ & $0.81 \pm 0.01$ & $0.90$ & $0.72$ & $0.90$\\
			\midrule
			Task 2 & MLP-L & $0.88$ & $0.91$ & $0.90 \pm 0.01$ & $0.84 \pm 0.01$ & $0.88$ & $0.34$ & $0.50$\\
			Task 2 & GRU-L & $\textbf{0.89}$ & $0.91$ & $\textbf{0.91} \pm 0.01$ & $0.84 \pm 0.01$ & $\textbf{0.89} \pm 0.01$ & $0.34$ & $0.50$\\
			\midrule
			Task 3 & MLP-L & $0.72 \pm 0.02$ & $0.99$ & $\textbf{0.88} \pm 0.01$ & $0.53 \pm 0.02$ & $0.50 \pm 0.05$ & $0.45$ & $0.50$\\
			Task 3 & GRU-L & $0.72 \pm 0.01$ & $0.99$ & $0.87 \pm 0.01$ & $0.53 \pm 0.02$ & $0.50 \pm 0.02$ & $0.45$ & $0.50$\\
			\midrule
			Task 4 & MLP-L & $0.69 \pm 0.02$ & $0.92$ & $0.84$ & $0.53 \pm 0.02$ & $0.47 \pm 0.04$ & $0.45$ & $0.50$\\
			Task 4 & GRU-L & $\textbf{0.70}$ & $0.92 \pm 0.01$ & $0.84 \pm 0.01$ & $0.53 \pm 0.01$ & $\textbf{0.50}$ & $0.45$ & $0.50$\\
			\midrule
			Task 5 & MLP-L & $\textbf{0.84} \pm 0.02$ & $0.97 \pm 0.01$ & $0.72 \pm 0.02$ & $\textbf{0.77} \pm 0.03$ & $\textbf{0.88} \pm 0.01$ & $0.41$ & $0.50$\\
			Task 5 & GRU-L & $0.83 \pm 0.01$ & $0.97$ & $0.72 \pm 0.02$ & $0.76 \pm 0.02$ & $0.87 \pm 0.01$ & $0.41$ & $0.50$\\
			\midrule
			Task 6 & MLP-L & $\textbf{0.85} \pm 0.01$ & $0.98$ & $\textbf{0.86} \pm 0.03$ & $0.78 \pm 0.03$ & $\textbf{0.77} \pm 0.06$ & $0.45$ & $0.50$\\
			Task 6 & GRU-L & $0.83 \pm 0.03$ & $0.98$ & $0.83 \pm 0.02$ & $0.78 \pm 0.02$ & $0.72 \pm 0.08$ & $0.45$ & $0.50$\\
			
			\bottomrule
		\end{tabular}
	}
	\caption{Aggregated test accuracies for each task (purely neural experiments, \textbf{Q1}). Values are reported as $mean \pm st. dev$, aggregated over 3 random seeds. Standard deviations are omitted if smaller than $10^{-2}$. MP: Most probable predictor.}
 \label{tab:neural-only}
\end{table*}

Figure \ref{fig:neural-only} and Table \ref{tab:neural-only} highlight the results of neural-only experiments.
In spite of the large training capacity, neural-only methods are characterized by slow convergence. In terms of average performance, only the simplest tasks (task 1, blue lines, and task 2, orange lines) achieve relatively high values by the end of training, with other tasks either plateauing at values close to random performance (tasks 3 and 4, green and red lines, respectively), or still far from convergence at the end of training (tasks 5 and 6, purple and brown). Some steep boosts in performance during training (e.g., tasks 2, between epochs 10-20, and 5, between epochs 20-40) might hint at optimization challenges.
\textsc{IC} performance is good for every task and it does not appear to be a bottleneck for performance. Tasks 3, 5 and 6 achieve near perfect classification accuracy, as they are characterized by values sampled from the MNIST digits domain, while the perceptually harder tasks 1, 2 and 4 are sampled from Fashion-MNIST, and still achieve $>0.9$ accuracy by the end of training.
In general there seem to be no difference between an \textsc{NSP} module implemented as multi-layer perceptron (MLP, solid lines) and one implemented with gated recurrent units (GRU, dotted lines). However, the harder-to-optimize tasks 2 and 5 present distinct training dynamics, with GRU converging in fewer epochs for task 2, while being slower for task 5. Task 2 is characterized by the necessity of ``keeping in memory'' a large temporal horizon (4 time steps, thus being easier for a recurrent architecture), albeit on perceptually simple constraints, while task 5 is a simple alternation of a complex (arithmetic) constraint being satisfied and then violated cyclically (thus architectural choice becomes irrelevant and feed-forward approaches, characterized by fewer learnable parameters, converge faster). There seem to be no backward interference of temporal modules on constraint and label accuracy, which are completely unaffected.
\textsc{CC} accuracy is generally good, with the exception of task 5, which cannot achieve satisfiable performance by the end of training. This task is characterized by a single arithmetic constraint in the form $A + B = C + D$, for which neural networks struggle. The initial random guessing phase corresponds to the first plateau in overall performance (epochs 0-20), as downstream objectives are conditioned on constraint prediction. Afterwards, in spite of low predictive performance at the \textsc{CC} level, the simple temporal behavior allows to achieve good \textsc{NSP} (and therefore average as well) accuracy in spite of an high rate of mispredicted constraints.
\textsc{NSP} is challenging for tasks 3 and 4, which do not improve over random guessing. Task 1 plateaus almost immediately, while tasks 2 and 5 are characterized by a slow start. Poor upstream predictive performance negatively affects task 5 at the beginning, while task 2 is subject to harder to optimize long-range temporal behavior.

\section{Auxiliary figures for Q3}
\begin{figure*}
	\centering
	\includegraphics[width=\linewidth]{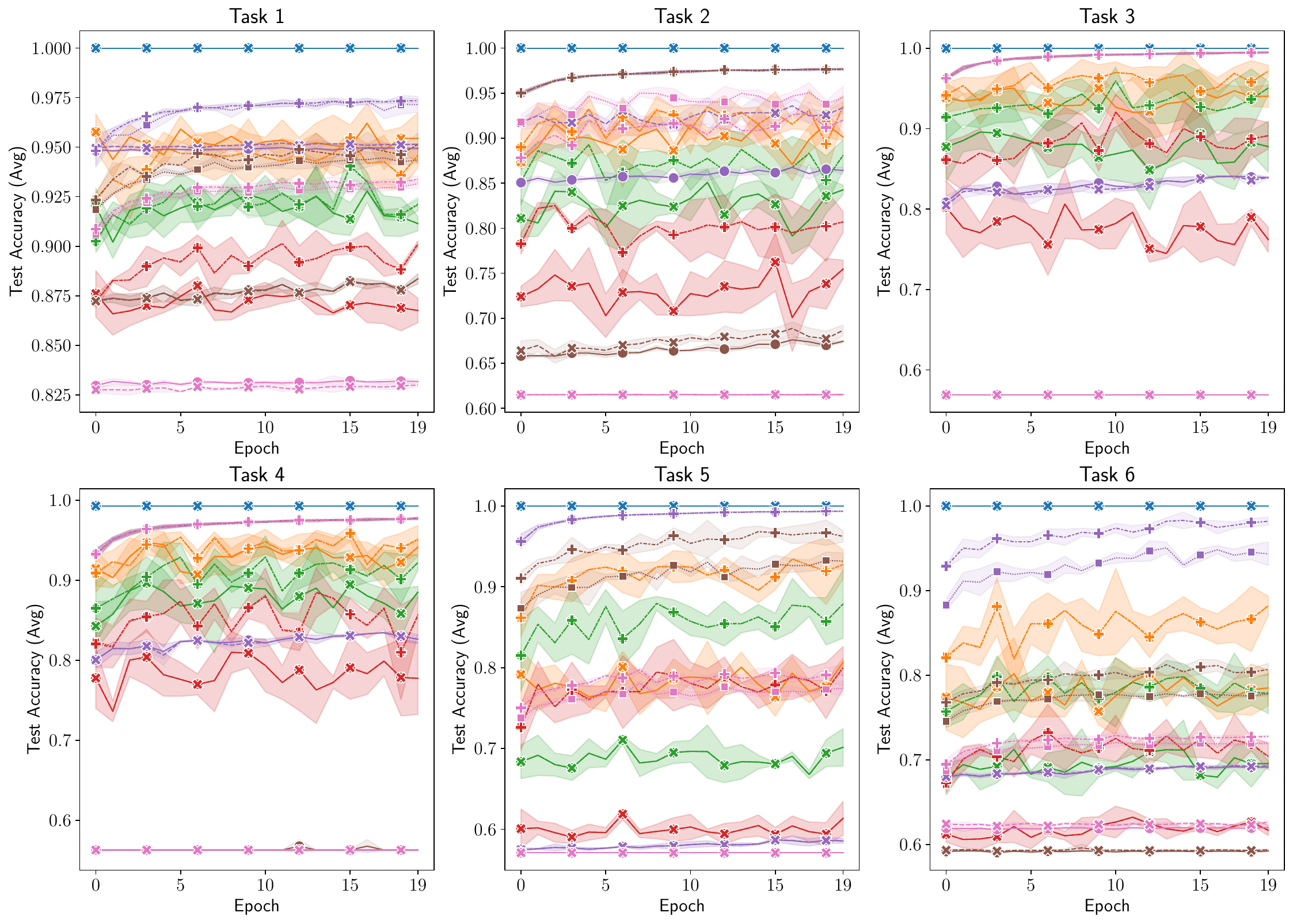}
    \includegraphics[width=.3\linewidth]{imgs/ablation_task4_legend.pdf}
	\caption{Average accuracy when replacing \textsc{IC} or \textsc{IC/CC} modules with an oracular predictor (\textbf{Q3}).}
	\label{fig:abl-avg}
\end{figure*}

\begin{figure*}
	\centering
	\includegraphics[width=\linewidth]{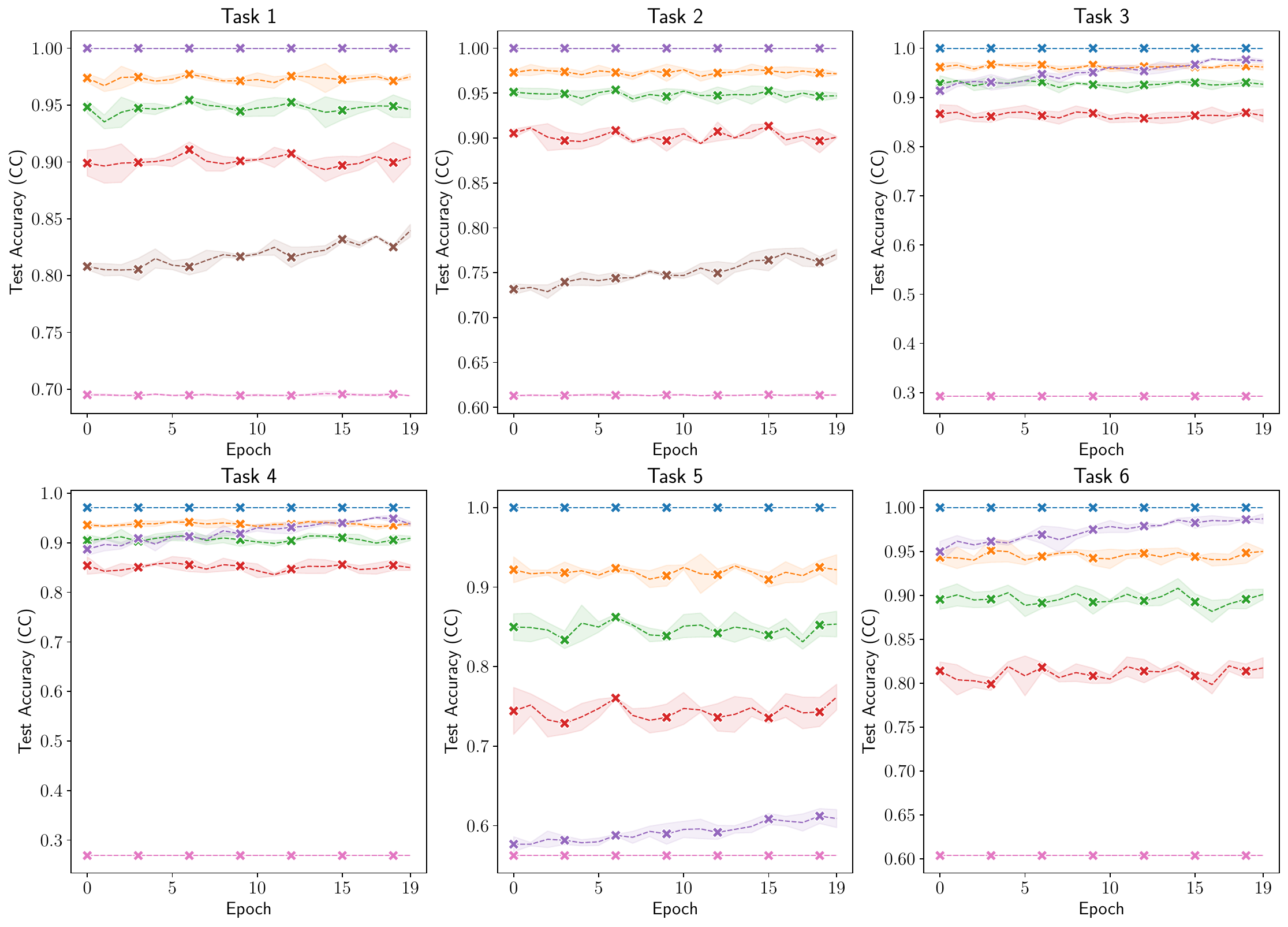}
    \includegraphics[width=.3\linewidth]{imgs/ablation_task4_legend.pdf}
	\caption{\textsc{CC} accuracy when replacing \textsc{IC} or \textsc{IC/CC} modules with an oracular predictor (\textbf{Q3}). For readability, \textsc{IC/CC} results are suppressed (they trivially achieve $1.0$ \textsc{CC} accuracy, as the argmax is the same as the perfect oracle). Only one \textsc{NSP} module is shown, as they do not affect \textsc{CC} performance.}
	\label{fig:abl-const}
\end{figure*}

\begin{figure*}
	\centering
	\includegraphics[width=\linewidth]{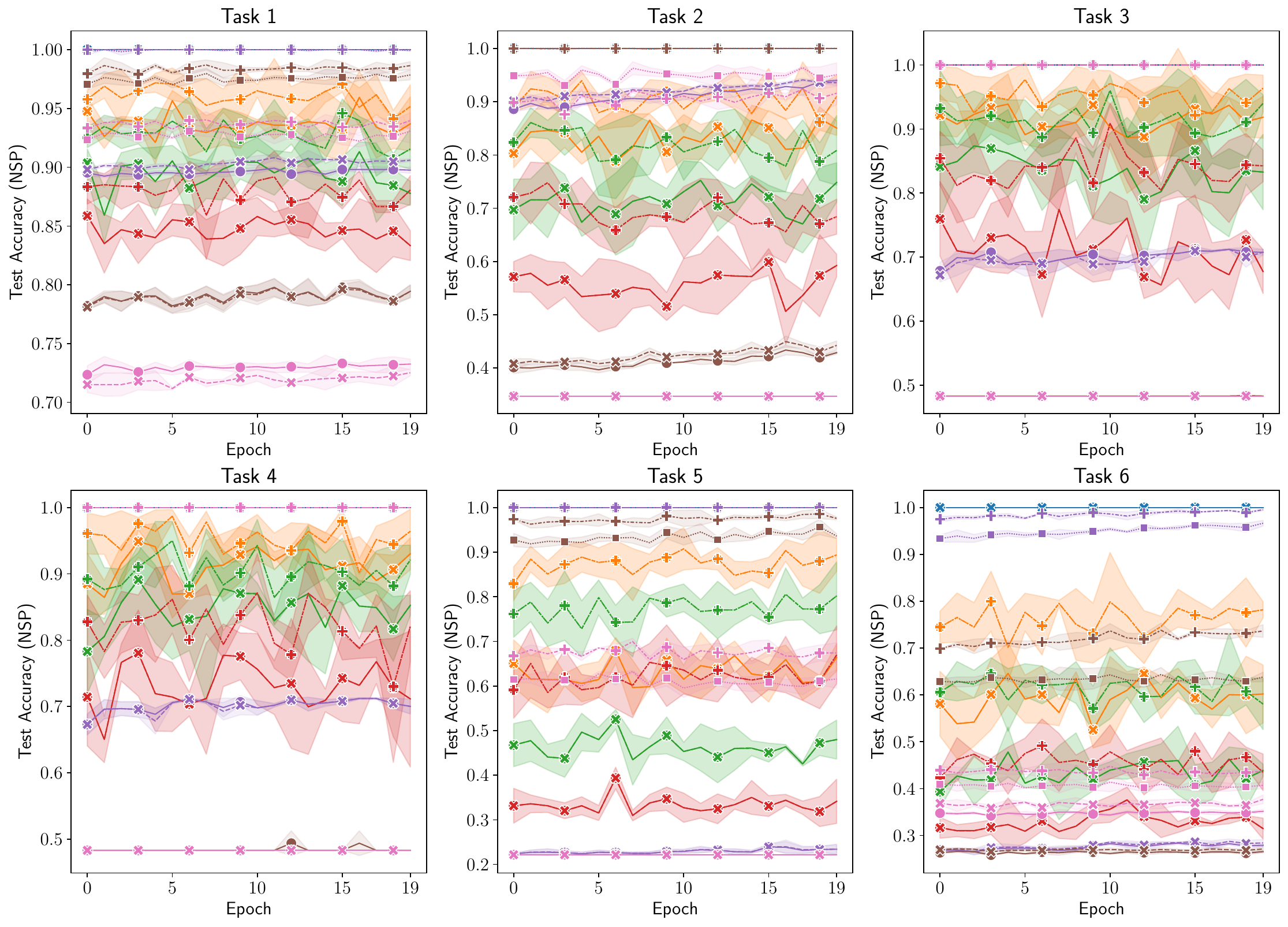}
    \includegraphics[width=.3\linewidth]{imgs/ablation_task4_legend.pdf}
	\caption{\textsc{NSP} accuracy when replacing \textsc{IC} or \textsc{IC/CC} modules with an oracular predictor (\textbf{Q3}).}
	\label{fig:abl-succ}
\end{figure*}

\begin{figure*}
	\centering
	\includegraphics[width=\linewidth]{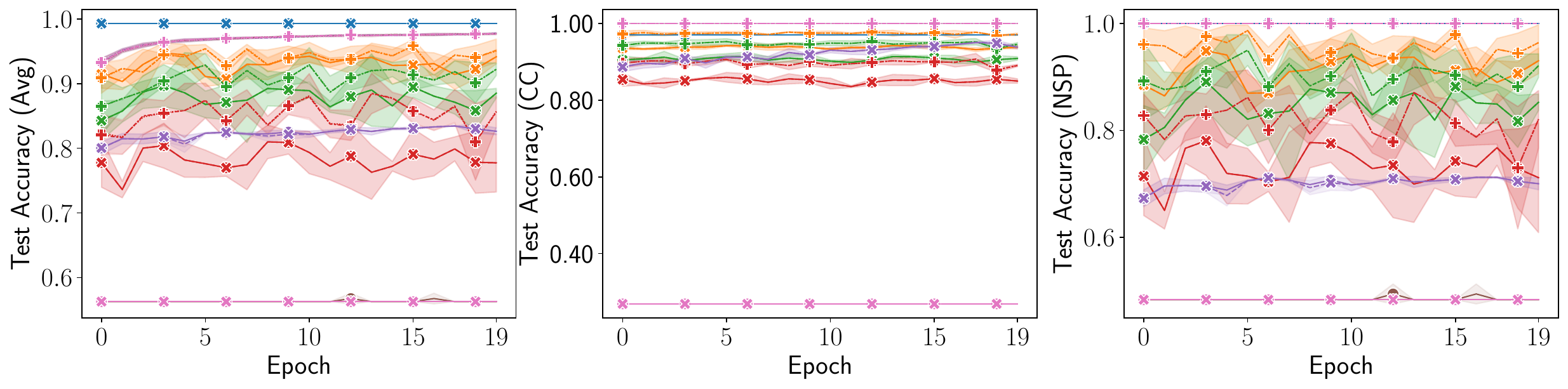}
    \includegraphics[width=.3\linewidth]{imgs/ablation_task4_legend.pdf}
	\caption{Accuracies for \textit{Task 4} with oracular predictors. These are the same data points as Fig. 2 in the main paper, showing also curves below threshold accuracy.}
	\label{fig:abl-task4-nolimits}
\end{figure*}

Figures \ref{fig:abl-avg}, \ref{fig:abl-const} and \ref{fig:abl-succ} show the behavior of symbolic methods fed with oracular predictors (Q3), for every task. For readability, Fig. \ref{fig:abl-task4-nolimits} shows accuracies for Task 4 side by side (this is the uncropped  version of Fig. 2 in the main paper).

\bibliographystyle{named}
\bibliography{biblio}